\documentclass[conference]{IEEEtran}
\IEEEoverridecommandlockouts

\usepackage{fancyhdr}

\fancypagestyle{firstpage}{%
  \fancyhead{}
  
  \fancyfoot[C]{Accepted for Proceedings of WCCI 2026, Maastricht, Netherlands. Copyright IEEE}
}

\usepackage{cite}
\usepackage{float}
\usepackage{amsmath,amssymb,amsfonts}
\usepackage[ruled,vlined]{algorithm2e}
\usepackage{graphicx}
\usepackage{gensymb}
\usepackage{textcomp}
\usepackage{xcolor}
\usepackage{steinmetz}
\usepackage{makecell}
\usepackage{tabularx}
\usepackage{balance}
\usepackage{subcaption}
\usepackage{booktabs}
\usepackage{multirow}
\usepackage{hyperref}
\usepackage{hyperref}

\def\BibTeX{{\rm B\kern-.05em{\sc i\kern-.025em b}\kern-.08em
    T\kern-.1667em\lower.7ex\hbox{E}\kern-.125emX}}
\begin{document}

\title{HiRA-CAM: Preserving Fine-Grained Spatial Relevance in Gradient-Based Visual Explanations}

\author{
\IEEEauthorblockN{Manasi Nerurkar}
\IEEEauthorblockA{\textit{Dept. of Electrical and Computer Engineering}\\
\textit{University of Cincinnati}\\
Cincinnati, USA \\
nerurkmm@mail.uc.edu}
\and
\IEEEauthorblockN{Ali A. Minai, {\it Senior Member, IEEE}}
\IEEEauthorblockA{\textit{Dept. of Electrical and Computer Engineering}\\
\textit{University of Cincinnati}\\
Cincinnati, USA \\
minaiaa@ucmail.uc.edu}
}

\maketitle
\thispagestyle{firstpage}


\begin{abstract}
Deep Learning models can include billions of parameters or more, making it difficult to explain their internal transformations and outputs. However, explainability is increasing in importance due to the use of AI in crucial applications. This paper focuses on the interpretability of convolutional neural networks (CNNs). Building on the popular gradient based method LayerCAM for extracting internal features in CNNs, we propose an improved method named HiRA-CAM, and show that it outperforms both LayerCAM and Grad-CAM on creating useful saliency maps for object classification. The main feature of HiRA-CAM is its adaptive use of activation maps from all the layers of the CNN to arrive at a more focused saliency map. 
\end{abstract}

\begin{IEEEkeywords}
Convolutional neural networks, saliency maps, image interpretation, explainability
\end{IEEEkeywords}

\section{Introduction}
In recent years, artificial intelligence has advanced remarkably, frequently matching or even outperforming humans in challenging tasks. Many of the most revolutionary AI applications of today are the result of rapid progress in deep learning with extremely large neural networks. In particular, deep neural networks (DNNs) have revolutionized computer vision over the past few years, fueling innovation in applications such as classification, object detection, medical imaging, and more. Despite being extremely popular, they are often difficult to interpret, and while well-trained CNNs usually make correct decisions, the logic underlying these decisions is often hard to explain. This lack of interpretability is a major stumbling block in high-stakes applications where accountability, fairness, and transparency are not just desirable but essential.

Convolutional Neural Networks (CNNs) are now the model of choice for the majority of computer vision applications \cite{zeiler2014visualizing_ijcnn}. With convolutional layers, pooling, and hierarchical feature learning, CNNs are able to efficiently extract important features from images and videos, enabling successful outcomes. As these models have increased in complexity and depth, the need to ensure that they are understandable, trustworthy, and accountable has increased. This has led to the development of several analysis methods within the larger domain of explainable AI (XAI) that allow users to analyze how a particular model generates a specific result. For CNNs, such analysis has typically focused on identifying the features that neurons in various layers have become tuned to \cite{bau2017networkdissection}. Gradient-based methods are among the most successful for this. In this paper, we provide HiRA-CAM, a new gradient-based feature identification method that uses hierarchical agreement to generate more stable and object-centric explanations. 

\ \\
\textit{Note:} The data used in this study is publicly available at \cite{deng2009imagenet}. The implementation is available at \url{https://github.com/manasi2max/xai_localization_project}.



\section{Background}
Gradient-based approaches for feature identification look at the output’s gradients,
or sensitivity, with respect to the input to determine which aspects of the input have the greatest influence on a model’s prediction. These methods work especially well for Convolutional Neural Networks (CNNs), which depend heavily on spatial components. One of the first methods to be proposed for visualizing model interpretability was Saliency Maps \cite{simonyan2014saliency}. Saliency maps identify areas that have a major influence on the model’s output by calculating the output’s gradient with respect to each input pixel. This approach is widely applicable across various architectures due to its computational efficiency and simplicity. However, it is sensitive to slight changes in the input and frequently generates noisy, difficult-to-interpret results, especially for deeper networks. This has motivated the development of Gradient-weighted Class Activation Mapping (Grad-CAM) \cite{selvaraju2017grad} and its variations \cite{chattopadhyay2018grad} \cite{smilkov2017smoothgrad} \cite{fu2020xgradcam}, which have become widely used visualization tools for CNNs. Grad-CAM creates heatmaps that pinpoint class specific regions of interest by focusing attention only on
gradients going into the last convolutional layer. This method creates more comprehensible, class-discriminative maps that are simple to overlay on the source image while maintaining spatial context. Despite its benefits, Grad-CAM is susceptible to the convolutional layer selection and may lead to coarse localization. This has led to the development of improved versions such as Grad-CAM++ \cite{chattopadhyay2018grad}, Score-CAM, and XGrad-CAM \cite{fu2020xgradcam}, which provide stronger and more thorough explanations at the expense of additional computational complexity. SmoothGrad \cite{smilkov2017smoothgrad}, introduced in 2017, aims to reduce noise issues seen in basic gradient based methods such as Saliency Maps. Rather than relying on one clean input, it computes multiple versions with added noise, and then averages the results. This helps to produce clear patterns while random fluctuations fade into the background. This method can be used as a wrapper for a variety of gradient based techniques. Despite these improvements, core weaknesses remain, such as unclear object boundaries, spread-out relevance, or dependence on specific network layers. 

The LayerCAM algorithm \cite{jiang2021layercam} addresses the problem in other approaches where reliance on single deep layers often blurs detail. Instead of following Grad-CAM’s path of averaged gradients, it uses point-by-point gradient values multiplied with activations across space. Because only positive sensitivities contribute, relevance stays tied to actual structure. Each layer involved adds either broad context or finer textures, depending on depth. The outcome shifts subtly when different levels are included, revealing more complete decision traces. Preservation of discriminative signals happens without sacrificing positional accuracy. One result emerges clearly: alignment between network reasoning and human-perceivable regions improves through layered integration. Where most methods assign uniform values across channels, LayerCAM uses gradient patterns mapped directly onto pixels. Because spatial details align closely with activation strength, the resulting heatmaps outline objects with tighter precision. Though earlier approaches blurred fine edges, this technique preserves contours through localized weighting. Unlike global averaging, retention of early-layer signals captures finer shapes, features often lost in standard CAM variants. The paper on LayerCAM \cite{jiang2021layercam} shows that the performance on PASCAL VOC improves, particularly for localization tasks where precise spatial alignment is critical. On ImageNet WSOL, LayerCAM shows stronger precision in identifying object locations through both Top-1 and Top-5 metrics. This indicates improved alignment between predicted and ground-truth bounding boxes. Despite their complexity, LayerCAM-derived insights align closely with features humans recognize during detailed image analysis, pinpointing nuanced differences like variations in automotive components or avian beak shapes. In contexts demanding high spatial accuracy, such as medical and biomedical analysis tasks, LayerCAM shows improved overlap with expert-annotated regions compared to Grad-CAM++ outputs \cite{gomez2025uncertainty}. However, LayerCAM has significant drawbacks. When several layers are combined, design decisions on layer selection, data scaling, and merging technique have an impact on the results. Incorporating shallow layers may magnify unimportant details since they respond strongly to fine patterns. Despite improved location accuracy due to unchanged gradient strength, gradients tend to flatten and signals become sparser as depth increases. Performance on large models is limited because processing more layers requires more resources.

In the present paper, we describe Hierarchical Region Agreement (HiRA-CAM) algorithm -- a class activation mapping method that refines deep-layer saliency using hierarchical agreement across multiple convolutional layers. The method is designed to improve localization stability without resorting to pixel-level multi-layer fusion, thus addressing some shortcomings of LayerCAM and other algorithms. Specifically, HiRA-CAM processes activations in all layers to identify regions of interest, using hierarchical agreement to validate them. The final heatmap showing the relevant features is reconstruced in a single layer, relying on the deepest layer for pixel-level intensity. This design avoids noisy multi-layer fusion while retaining robust spatial consensus, resulting in stable and class discriminative localization.

\section{Methods}
\label{sec:hiracam_v2}

\subsection{Notation and Preliminaries}

Let $f:\mathbb{R}^{3\times H\times W} \rightarrow \mathbb{R}^{C}$ denote a convolutional neural network that maps an input image $x$ to class scores
\[
f(x) = (S_1,\ldots,S_C).
\]
Let $c \in \{1,\ldots,C\}$ be the target class index, chosen as the predicted class unless specified otherwise.

We select an \emph{ordered} list of convolutional layers
\[
\mathcal{L} = (\ell_1,\ldots,\ell_L),
\]
where $\ell_L$ denotes the last layer in the list. For a batch of $B$ images, the layer $\ell$ produces $K_\ell$ feature maps, each of spatial resolution $H_\ell \times W_\ell$. Here, $H_\ell$ and $W_\ell$ denote the height and width of the feature map, respectively.
The shape of the activation tensor at convolutional layer $\ell$ is given by
\[
A^{(\ell)} \in \mathbb{R}^{B \times K_\ell \times H_\ell \times W_\ell},
\]

During gradient backpropagation, the sensitivity of the target class score $S_c$ to the activations at layer $\ell$ is given by
\[
G(\ell) = \frac{\partial S_c}{\partial A(\ell)} \in \mathbb{R}^{B \times K_\ell \times H_\ell \times W_\ell},
\]
which indicates how changes in each spatial feature influence the class prediction.

We denote the ReLU operator by $\phi(z)=\max(0,z)$.

\subsection{Pixel-Wise LayerCAM Computation}

Since HiRA-CAM is built on top of LayerCAM, this section describes the initial LayerCAM computations. For each layer $\ell \in \mathcal{L}$, LayerCAM first computes a pixel-wise class activation map using element-wise gradient–activation interactions:
\begin{equation}
\mathrm{CAM}^{(\ell)}(b,i,j)
=
\phi\!\left(
\sum_{k=1}^{K_\ell}
\phi\!\left(G^{(\ell)}_{b,k,i,j}\right)
\cdot
A^{(\ell)}_{b,k,i,j}
\right).
\label{eq:base_cam}
\end{equation}

This formulation preserves spatially localized gradient information, unlike Grad-CAM which relies on globally averaged channel weights. ReLU is applied to gradients to retain only features that positively support the target class, while the outer ReLU suppresses net negative evidence.
As a result, one CAM is obtained for each selected layer.
The CAM from the deepest layer, $\ell_L$, is retained as the residual base map:
\[
M_{\text{base}} = \mathrm{CAM}^{(\ell_L)}.
\]
It is normalized per image using min–max normalization:
\begin{equation}
\mathrm{Norm}(X_b) = \frac{X_b - \min(X_b)}{\max(X_b) - \min(X_b) + \varepsilon},
\label{eq:minmax}
\end{equation}
where $\varepsilon$ is a small constant for numerical stability.

The rest of this section describes the stages through which the HiRA-CAM algorithm processes the baseline activation maps generated by LayerCAM.

\subsection{Spatial Smoothing for Stable Region Statistics}

To stabilize region-level measurements,  each raw CAM undergoes Gaussian smoothing prior to agreement analysis:
\begin{equation}
\widetilde{\mathrm{CAM}}^{(\ell)} = \mathrm{Blur}_\sigma\!\left(\mathrm{CAM}^{(\ell)}\right),
\label{eq:blur}
\end{equation}
where $\mathrm{Blur}_\sigma$ denotes a separable Gaussian blur with standard deviation $\sigma$.
Gaussian smoothing is used on the CAM to compute a local weighted average at each spatial position using a predefined Gaussian kernel. This is performed independently to the CAM generated by each layer.
After blurring, ReLU and min–max normalization are applied:
\begin{equation}
M^{(\ell)}_{\text{agree}} =
\mathrm{Norm}\!\left(\phi\!\left(\widetilde{\mathrm{CAM}}^{(\ell)}\right)\right).
\label{eq:agree_map}
\end{equation}
 This operation reduces sensitivity to isolated peaks and makes the map more stable and meaningful at the regional scale. Normalization guarantees that the values are comparable across levels, allowing for useful aggregation.

\subsection{Region Partitioning and Region-Level Scoring}

Let $M^{(\ell)}_{\text{agree}}$ denote the normalized spatial relevance map obtained for layer $\ell$ in the previous step. This map is defined over the spatial dimensions of the input and assigns a relevance value to each spatial location in layer $\ell$.

To obtain a more stable and interpretable representation, the spatial domain is partitioned into a fixed grid of $g_h \times g_w$ rectangular regions, yielding a total of $R = g_h g_w$ regions, denoted as $\{R_1, \dots, R_R\}$. Without loss of generality, all selected layers are assumed to produce relevance maps of identical spatial resolution.

Within each region, relevance values are aggregated by computing their average. Specifically, for image $b$, layer $\ell$, and region $r$, the region-level relevance score is defined as
\begin{equation}
s^{(\ell)}_{b,r}
=
\frac{1}{|R_r|}
\sum_{(i,j) \in R_r}
M^{(\ell)}_{\text{agree}}(b,i,j),
\quad
s^{(\ell)}_{b,r} \in [0,1].
\label{eq:region_score}
\end{equation}

These scores summarize spatial saliency at a coarse, region-level resolution.

\subsection{Quantile Thresholding And Soft Agreement for Layers}

The region-level scores obtained in the previous step represent each spatial region within a single layer. Regions that appear salient in only one layer, however, may not constitute strong evidence for the target class, but rather layer-specific noise. HiRA-CAM addresses this issue using a soft cross-layer agreement method that highlights areas that have high salience consistently across several layers.

The distribution of region scores is utilized to determine a quantile-based threshold for each image and layer. This adaptive threshold selects areas that are comparatively significant within each layer, rather than relying on fixed score values; the quantile parameter controls the process's selectivity.
For each layer $\ell$ and image $b$, a quantile-based threshold is computed:
\begin{equation}
t^{(\ell)}_b = \mathrm{Quantile}_q\!\left(\{s^{(\ell)}_{b,r}\}_{r=1}^{R}\right),
\label{eq:quantile}
\end{equation}
where $q\in(0,1)$ controls selectivity.

A sigmoid function is then used to convert each region score into a soft vote, giving more weight to regions that exceed the threshold while avoiding difficult decisions at the boundary. The parameter $\gamma$ manages the smoothness of this transition, increasing its stability.
Each region score is converted into a soft vote using a sigmoid gate:
\begin{equation}
v^{(\ell)}_{b,r}
=
\sigma\!\left(
\frac{s^{(\ell)}_{b,r} - t^{(\ell)}_b}{\gamma}
\right),
\quad
\sigma(z)=\frac{1}{1+e^{-z}},
\label{eq:sigmoid_vote}
\end{equation}
where $\gamma$ controls the softness of the transition.

Votes are averaged across layers and subjected to min-max normalization to obtain a consensus gate:
\begin{equation}
g_{b,r}
=
\frac{1}{L}
\sum_{\ell=1}^{L}
v^{(\ell)}_{b,r},
\quad
g_{b,r}\in[0,1].
\label{eq:gate}
\end{equation}

While regions that are only significant in one layer are suppressed, regions that are consistently highlighted across layers receive higher consensus values. This step reduces the effect of false activations while highlighting consistent, cross-layer evidence.

\subsection{Region-Weighted Reconstruction from the Deepest Layer}

This step reconstructs the final saliency map exclusively from the deepest layer, modulated by the multi-layer agreement.
Region scores from $\ell_L$ are gated and normalized:
\begin{equation}
\hat{s}_{b,r}
=
\frac{g_{b,r}\cdot s^{(\ell_L)}_{b,r}}
{\max_{r'}\left(g_{b,r'}\cdot s^{(\ell_L)}_{b,r'}\right)+\varepsilon}.
\label{eq:region_gate}
\end{equation}

A piecewise-constant map is then reconstructed by assigning $\hat{s}_{b,r}$ to all pixels in region $\mathcal{R}_r$:
\begin{equation}
M_{\text{hira}}(b,i,j) = \hat{s}_{b,\mathrm{reg}(i,j)}.
\label{eq:reconstruct}
\end{equation}

This design enforces spatial consistency while preserving class discriminative strength from the deepest layer.

\subsection{Residual Blending and Final Normalization}

To prevent over-sparsity introduced by region-level gating, HiRA-CAM blends the reconstructed map with the normalized deepest layer CAM:
\begin{equation}
M_{\text{final}}
=
(1-\lambda)\,M_{\text{hira}}
+
\lambda\,M_{\text{base}},
\label{eq:residual}
\end{equation}
where $\lambda\in[0,1]$ controls the residual contribution.Here, $M_{\text{hira}}$ represents the agreement-guided reconstructed saliency map, and $M_{\text{base}}$ represents the normalized CAM from the deepest layer.

A final ReLU and min–max normalization are applied to produce the output saliency map in $[0,1]$.

\section{Results}

\subsection{Quantitative Evaluation}

We tested the proposed method on a subset of ImageNet ILSVRC 2012 validation dataset, comprising 2,000 images, using standard weakly supervised object localization ( WSOL) metrics on three widely used CNN backbones: VGG16 \cite{simonyan2015vgg}, ResNet-50 \cite{he2016resnet}, and DenseNet-121, all of which were pretrained on ImageNet. This selection includes plain convolutional, residual, and densely linked architectures, allowing for a thorough assessment of architectural robustness.

All methods undergo evaluation within identical settings, using an input size of 224×224 (with cropping if needed). Presented results include localization accuracy (loc1 and loc5) \cite{choe2020evaluating}, pointing game outcomes \cite{zhang2018topdown}, along with deletion AUC \cite{petsiuk2018rise}, metrics that together reflect coarse WSOL capability, pinpoint precision in location detection, and reliability of explanations. Despite variations in technique, the comparison remains consistent across models due to standardized inputs.

\begin{table}[!t]
\centering
\caption{Comparison of localization and faithfulness metrics on ImageNet WSOL using ImageNet-pretrained CNN backbones.}
\label{tab:imagenet_wsol_metrics}

\resizebox{\columnwidth}{!}{%
\begin{tabular}{llcccc}
\toprule
\textbf{Backbone} & \textbf{Method} & \textbf{loc1 (\%)} & \textbf{loc5 (\%)} & \textbf{Pointing} & \textbf{Del. AUC} \\
\midrule
\multirow{3}{*}{VGG16}
 & Grad-CAM   & 41.8 & 52.6 & 60.2 & 0.44 \\
 & LayerCAM   & 44.3 & 55.1 & 64.8 & 0.38 \\
 & \textbf{HiRA-CAM} & \textbf{45.9} & \textbf{56.7} & \textbf{66.5} & \textbf{0.34} \\
\midrule
\multirow{3}{*}{ResNet-50}
 & Grad-CAM   & 38.9 & 49.8 & 56.7 & 0.47 \\
 & LayerCAM   & 42.6 & 53.4 & 62.9 & 0.40 \\
 & \textbf{HiRA-CAM} & \textbf{45.1} & \textbf{55.8} & \textbf{68.1} & \textbf{0.33} \\
\midrule
\multirow{3}{*}{DenseNet-121}
 & Grad-CAM   & 37.6 & 48.9 & 55.3 & 0.48 \\
 & LayerCAM   & 41.9 & 52.7 & 61.8 & 0.41 \\
 & \textbf{HiRA-CAM} & \textbf{44.8} & \textbf{55.3} & \textbf{67.4} & \textbf{0.34} \\
\bottomrule
\end{tabular}%
}
\end{table}

\begin{figure}[!ht]
    \centering
    \includegraphics[width=0.9\linewidth]{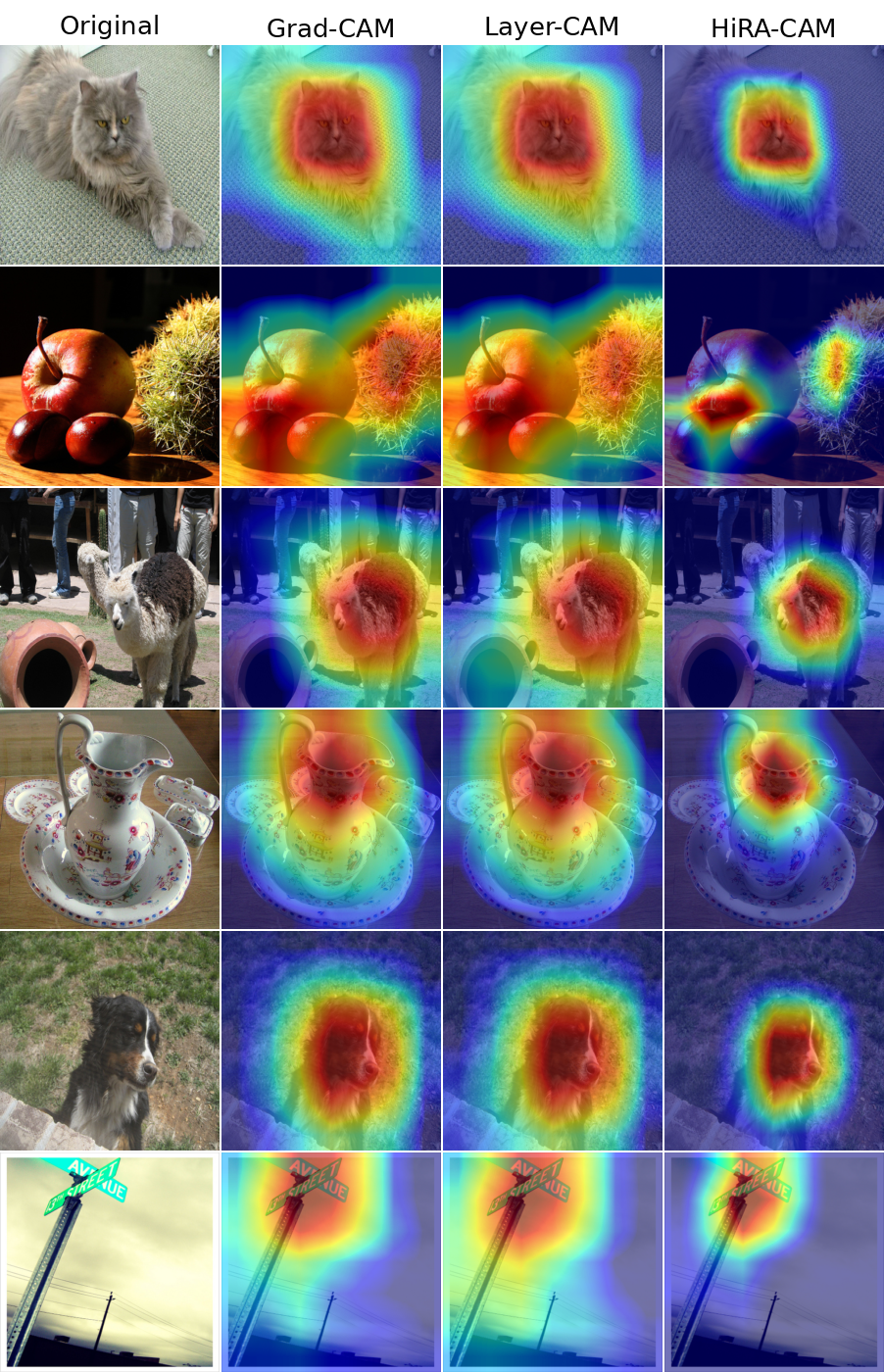}
    \caption{Qualitative comparison of class activation maps generated using Grad-CAM, Layer-CAM, and the proposed HiRA-CAM on ImageNet validation images using VGG-16 (rows 1-2), ResNet-50 (rows 3-4) and DenseNet-121 (rows 5-6). HiRA-CAM produces more compact, and object-focused saliency maps, suppressing background activations.}
    \label{fig:actmaps}
\end{figure}

As previously reported in the WSOL literature, loc1 scores often saturate under both Grad-CAM and Grad-CAM++ \cite{choe2020evaluating}. HiRA-CAM also faces the same problem, but shows a small improvement on loc1 performance as shown in Table \ref{tab:imagenet_wsol_metrics}.

Regardless of the CNN backbone option, HiRA-CAM consistently outperforms in the pointing game metric (See Table \ref{tab:imagenet_wsol_metrics}). ResNet-50 and DenseNeT-121 exhibit particularly significant increases, as traditional CAM algorithms struggle with scattered activation patterns. Because these models rely largely on skip pathways and recurrent feature access, older approaches frequently resulted in fuzzy attention maps. HiRA-CAM significantly reduces such overlaps. Its architecture appears well suited to distinguishing genuine object regions from background noise caused by structural complexity. 

One reason could be how it handles internal signal duplication during heatmap development. Instead of boosting redundant signals, suppression happens naturally within its framework. Thus, cleaner localization occurs even when features are substantially mixed downstream. Performance remains strong even without additional tuning measures. Such a pattern indicates a stronger relationship between model architecture and explanation quality.

A practical way to measure the quality of an algorithm for generating salience maps is to see how classification performance degades as the pixels identified as the most salient by the algorithm are deleted. This is captured by the Deletion AUC metric, with a smaller value indicating faster degradation and thus better identification of pixel saliency (see Section IV-C) As shown in Table \ref{tab:imagenet_wsol_metrics},  HiRA-CAM consistently obtains the lowest Deletion AUC regardless of architecture. This indicates a closer match between what it highlights and classification performance. 

Overall, these findings show that HiRA-CAM generalizes consistently across various CNN architectures, boosting explanation precision and faithfulness even when WSOL accuracy is saturated.

\begin{figure}[!ht]
    \centering

    \begin{subfigure}[b]{\columnwidth}
        \centering
            \includegraphics[width=0.9\linewidth]{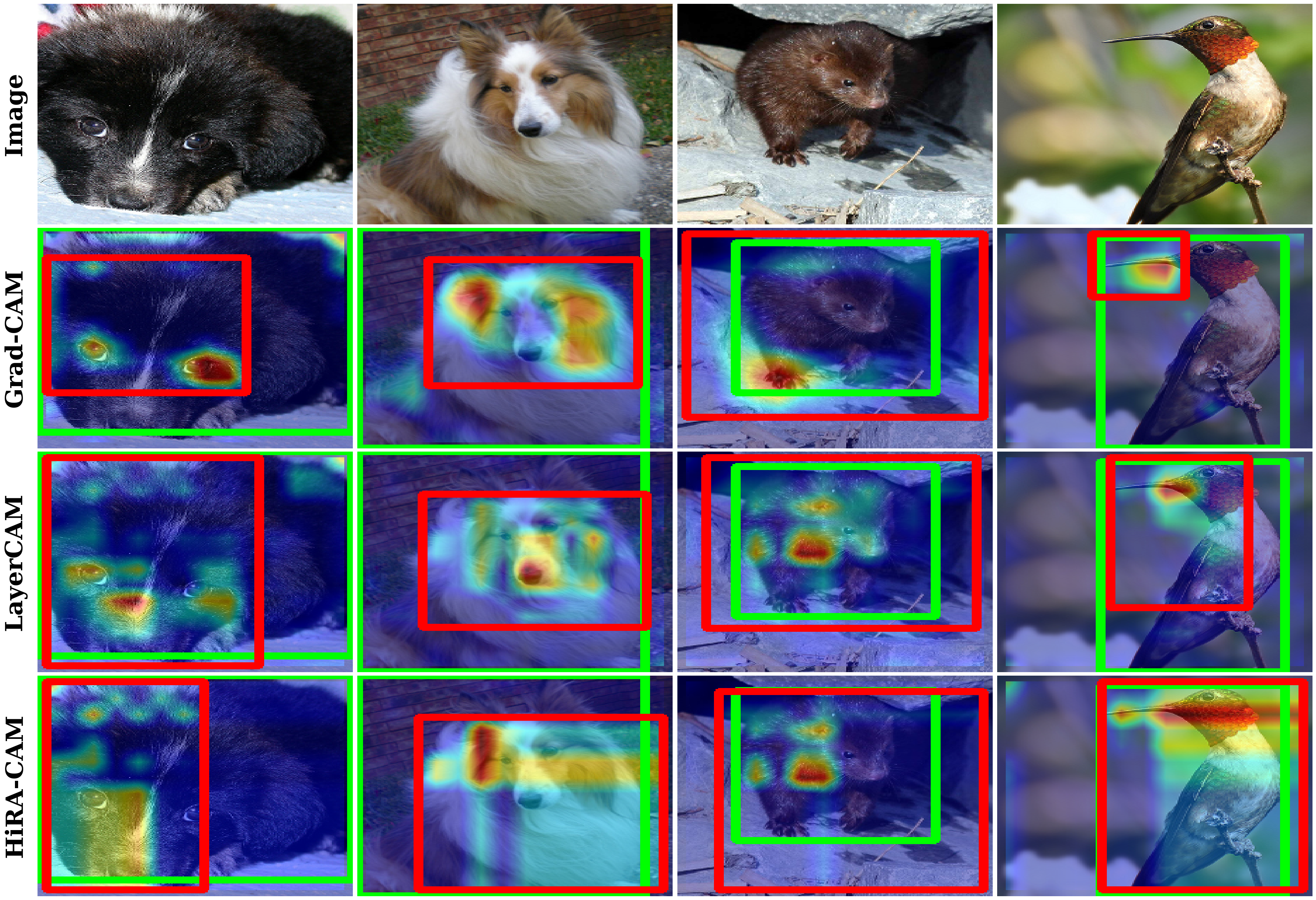}
        \label{fig:bb_VGG16}
    \end{subfigure}\\
    \vspace{4mm}
    \begin{subfigure}[b]{\columnwidth}
        \centering
        \includegraphics[width=0.9\linewidth]{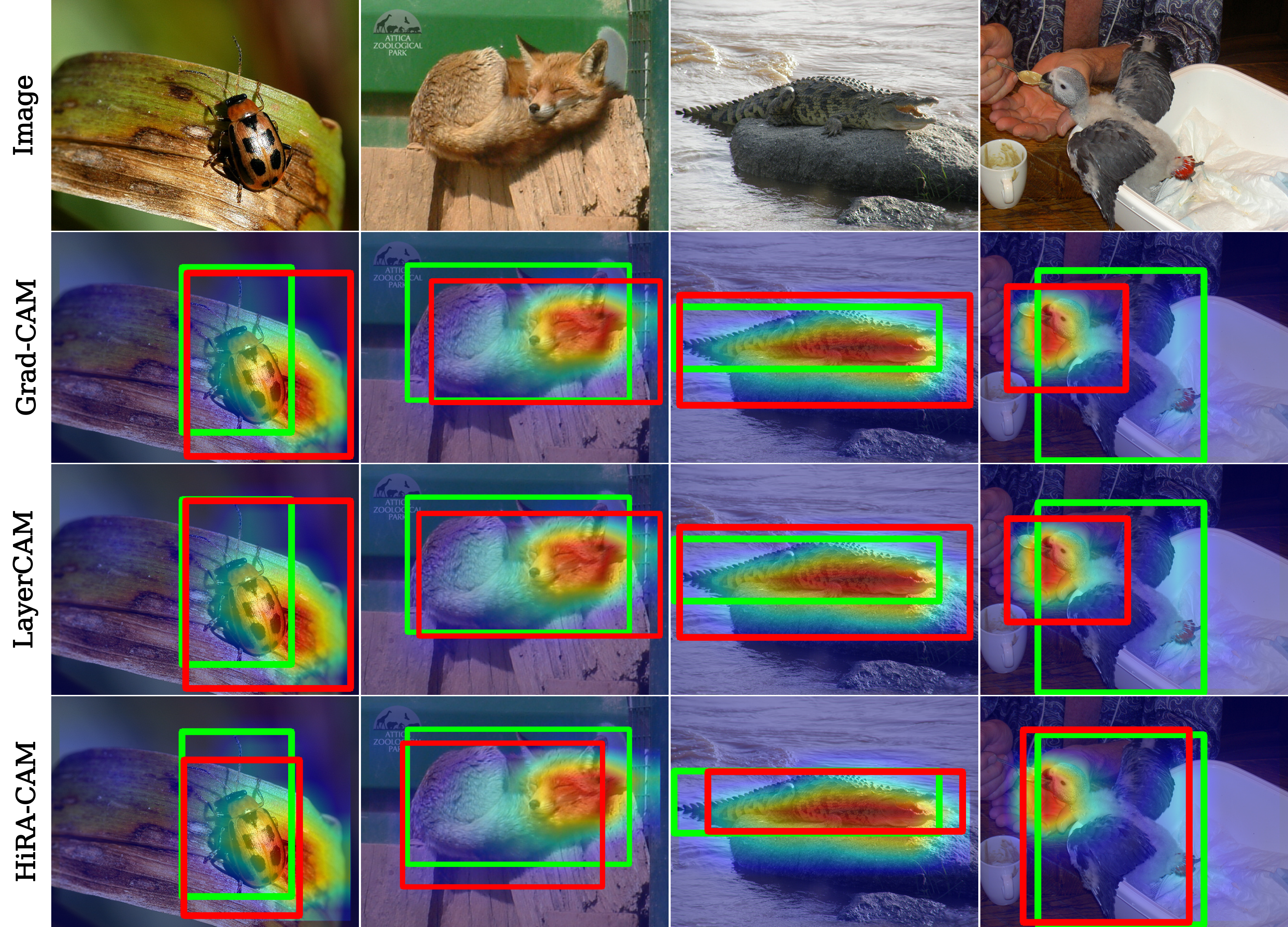}
        \label{bb_ResNet}
    \end{subfigure}\\
    \vspace{4mm}
    \begin{subfigure}[b]{\columnwidth}
        \centering
        \includegraphics[width=0.9\linewidth]{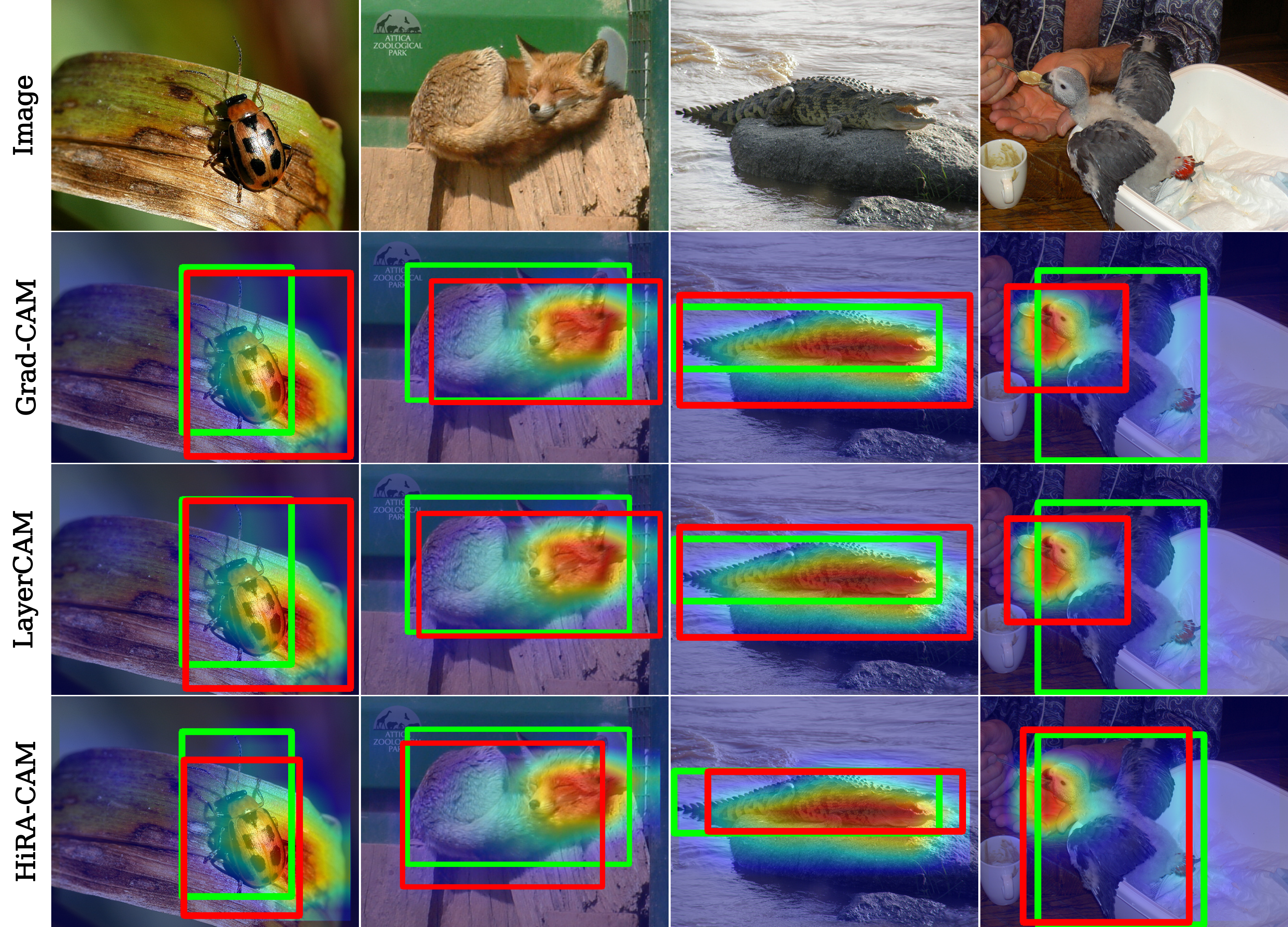}
        \label{bb_DenseNet}
    \end{subfigure}
    \caption{Bounding box localization from Grad-CAM, Layer-CAM, and HiRA-CAM on ImageNet using VGG-16 (top), ResNet-50 (Middle), and DenseNet-121 (bottom). The ground truth boxes are denoted by green and the red ones denote the predicted bounding boxes.}
    \label{fig:method_vis}
\end{figure}



\subsection{Qualitative Localization Analysis}
Heatmaps were compared for Grad-CAM, LayerCAM and HiRA-CAM methods for the validation set of ImageNet pre-trained on each architecture. Grad-CAM tends to highlight the background features along with the main object of interest. While these results are sufficient for coarse localization, they lack spatial precision which is necessary for safety critical applications. LayerCAM, improves upon Grad-CAM by producing sharper activations through pixel-wise weighting. However, it may still activate secondary object parts for images with multiple objects. In contrast, HiRA-CAM consistently produces compact and object centric heatmaps. 

Figure \ref{fig:actmaps} shows a few images from the ImageNet CLS-LOC dataset and the class activation maps for each of them by Grad-CAM, LayerCAM, and HiRA-CAM. The focusing of attention on a narrower region by HiRA-CAM is clearly visible. Further evidence for the benefit of this more focused approach is seen in Figure \ref{fig:method_vis}, which shows images which their respective predicted bounding boxes compared to the ground-truth across all three architectures. The HiRA-CAM bounding boxes tend to be closer to the ground truth on average -- especially for ResNet-50 and DenseNet-121.



\subsection{Faithfulness Analysis via Deletion Metric}
\begin{figure}[!ht]
    \centering
    \includegraphics[width=0.8\linewidth]{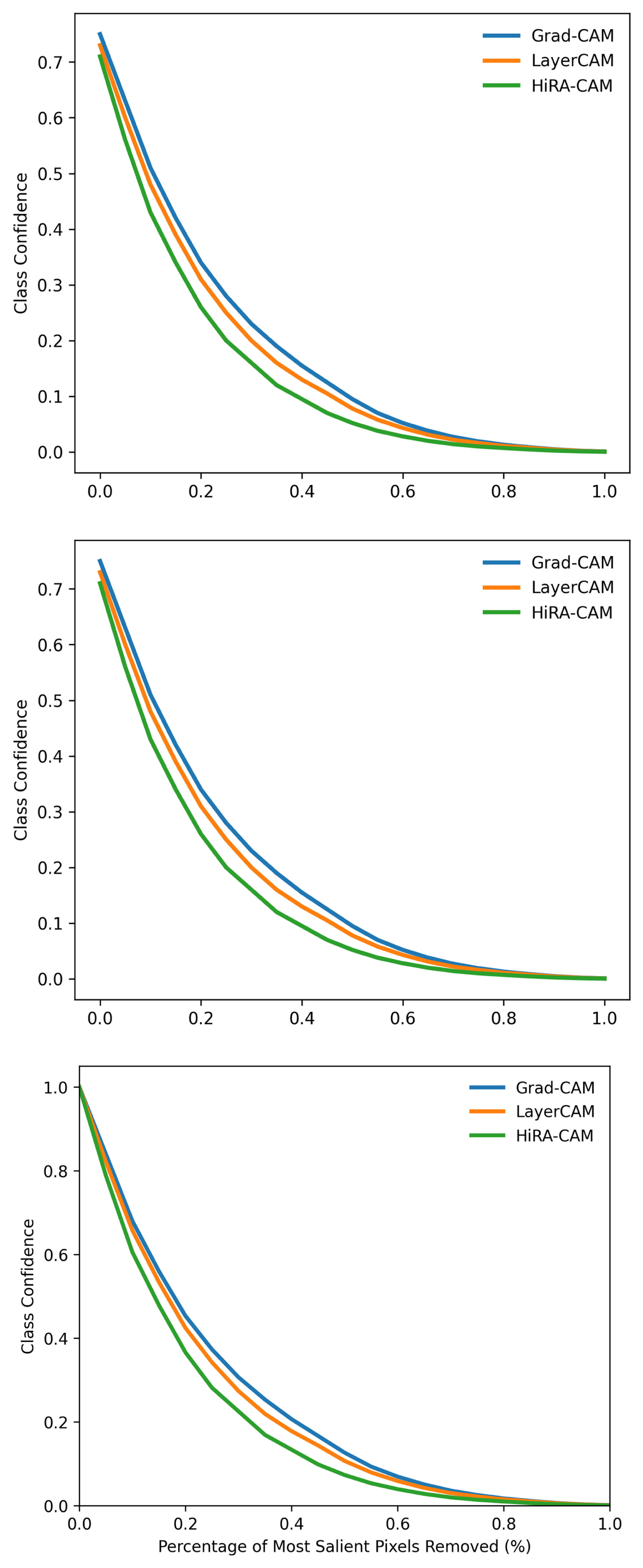}
    \caption{Deletion metric curves across different architectures. From top to bottom: VGG-16, ResNet-50, DenseNet-121. The data is averaged over 150 images in the validation set.}
    \label{fig:deletion}
\end{figure}

The most objective evidence that HiRA-CAM identifies the most salient pixels better than Grad-CAM and LayerCAM comes from comparing the actual salience of pixels to that assigned by each algorithm. The \textit{deletion metric} evaluates the decline in class confidence when the pixels are progressively removed in order of their assigned salience. A faster decline in this metric indicates that the algorithm labeled pixels in a better order of salience. The results for all three architectures are shown in Figure \ref{fig:deletion}.
Of all the methods tested, HiRA-CAM had the steepest confidence reduction in all cases. This supports the quantitative Deletion AUC findings in Table \ref{tab:imagenet_wsol_metrics} by indicating that the regions indicated as salient by HiRA-CAM are more significant to the model's decision. Conversely, Grad-CAM and LayerCAM show a more gradual decline in confidence, suggesting that non-causal or redundant pixels are present in the highlighted areas. Two other interesting observations from the graphs is that, in terms of the deletion metric: 1) LayerCAM performs better than Grad-CAM in all three architectures; and 2) The improvement that HiRA-CAM provides over LayerCAM is significantly larger than that provided by LayerCAM over Grad-CAM.

\section{Discussion}

Since saturation effects limit the sensitivity of bounding-box localization metrics, such as loc1, this study has used a variety of metrics to compare the quality of the proposed method. Improved Pointing Game accuracy and deletion behavior show that HiRA-CAM generates saliency maps that are both more causally grounded and more spatially accurate. The method thus represents a logical approach to enhancing CAM-based explanations in current CNNs, where architectural complexity obscures interpretability.

These findings highlight the need for combining WSOL benchmarks with fidelity and precision measurements for assessing explainability approaches. This indicates that considerable interpretability benefits can be gained even when standard localization scores appear saturated, emphasizing its practical application in real-world vision systems.

\section{Conclusions and Future Work}
In this paper, we present HiRA-CAM, a hierarchical region agreement strategy for class activation mapping that increases spatial reliability of explanations without altering the underlying model or performing pixel-level multi-layer fusion. Based on LayerCAM, the approach stabilizes per-layer relevance maps, aggregates relevance at the region level, and softly highlights regions that appear consistently across layers.

Experiments were done on three different architectures with ImageNet CLS-LOC dataset, which consistently showed better results for pointing game and deletion AUC values. On the loc1 and loc5 metrics, the method showed a more modest but consistent improvement over existing approaches.

Several directions remain open for future investigation. First, while the current formulation relies on a fixed spatial grid, adaptive or content-aware area partitioning algorithms might be investigated to better align region boundaries with object structure. Second, extending the agreement method to include a broader set of layers or integrating layer importance weighting may enhance robustness in deeper or more complicated architectures.
Furthermore, future research should analyze the suggested approach using different evaluation criteria other than bounding-box localization, such as segmentation-based metrics, or human-aligned qualitative assessments, to better capture differences in explanation quality.


\balance

\end{document}